\documentclass[cameraready]{Interspeech}
\usepackage{multirow}
\usepackage{threeparttable}
\usepackage{booktabs}
\usepackage{supertabular}
\usepackage{longtable}
\usepackage{multicol}
\usepackage{subcaption}
\usepackage{array}
\usepackage{pifont}
\usepackage[table]{xcolor}

\title{UR-BERT: Scaling Text Encoders for Massively Multilingual TTS\\Through Universal Romanization and Speech Token Prediction}

\author[affiliation={}]{Sangmin}{Lee}
\author[affiliation={}]{Eekgyun}{Ahn}
\author[affiliation={}]{Woongjib}{Choi}
\author[affiliation={}]{Hong-Goo}{Kang}

\address{
    Dept. of Electrical and Electronic Engineering, Yonsei University, Seoul, South Korea
}

\email{\{sangmin\_lee, woongzip1\}@dsp.yonsei.ac.kr, kennyahn0830@gmail.com, hgkang@yonsei.ac.kr}


\usepackage{comment}
\usepackage{cite}
\usepackage{hyperref}

\begin{document}

\maketitle

\begin{abstract}
We propose UR-BERT, a Romanized transcription-based text-to-speech (TTS) encoder for massively multilingual TTS systems. Conventional grapheme-to-phoneme (G2P)-based approaches are limited to around 100 languages due to the availability of reliable G2P resources. In contrast, UR-BERT scales to 495 languages by unifying diverse writing systems into a shared Romanization representation.
To further enhance phonetic fidelity and text–speech alignment, we introduce a speech token prediction objective during training, which encourages the encoder to learn speech-aware phonetic representations in a data-efficient manner.
Experiments show that TTS systems built on UR-BERT consistently outperform recent text encoder baselines across a wide range of languages and resource conditions, and demonstrate strong generalization to unseen languages.
\end{abstract}

\section{Introduction}
\label{intro}
Neural text-to-speech (TTS) systems have achieved substantial progress across languages and speaking styles. Most recent approaches adopt encoder–decoder architectures, in which the encoder produces linguistic representations that are transformed into acoustic features or speech waveforms by a decoder. While decoder models have advanced rapidly with the introduction of flow matching and neural codec language modeling~\cite{tacotron,fastspeech2,glowtts,vits,gradtts,difftts,matchatts,f5tts,valle,speartts}, encoder design has received comparatively less attention. In particular, prior work primarily focused on phonetic adequacy for reliable text–speech alignment, a recurring challenge in TTS.

Meanwhile, advances in self-supervised learning have demonstrated strong empirical performance across diverse domains~\cite{bert, albert, roberta, wav2vec2, hubert, wavlm}, fostering growing interest in the pretraining of text encoders for TTS. These models capture rich contextual and semantic information beyond purely phonetic cues, leading recent TTS systems to incorporate BERT-style representations to enhance naturalness. 
Prior studies~\cite{ebert1,ebert2,ebert3,ebert4} incorporated BERT embeddings as auxiliary inputs to augment phonetic representations. However, this approach exposes a structural mismatch between TTS text encoders and general-purpose language models. Specifically, TTS systems typically operate at the character- or phoneme-level, whereas BERT relies on subword units, creating a granularity discrepancy that complicates precise alignment and representation integration.

To mitigate this mismatch, subsequent work proposed BERT-style text encoders pretrained from scratch to better align linguistic representations with TTS requirements. Early approaches~\cite{pngbert,mpbert} introduced phoneme-aware pretraining as a core design principle by jointly modeling grapheme and phoneme units to bridge textual and phonetic spaces. Building on this paradigm, a later extension~\cite{plbert} streamlined the framework by incorporating both grapheme and phoneme information only during pretraining while restricting downstream usage to phoneme inputs. More recently, studies~\cite{styletts2,xphonebert} have extended phoneme-level pretraining to multilingual settings, demonstrating that phoneme-based language modeling remains effective even when trained on multilingual corpora.

Despite their effectiveness, these models inherently rely on G2P toolkits~\cite{phonemizer,charsiug2p} to generate phoneme sequences, creating a systemic dependency that significantly constrains scalability. This reliance poses a major obstacle to achieving truly global coverage, as G2P systems are typically available for only around 100 languages, leaving the vast majority of the world’s languages unsupported. In addition, encoders pretrained solely on textual corpora lack exposure to acoustic contexts, preventing them from capturing fine-grained prosodic and speech-related cues that are critical for high-quality TTS synthesis.

To address these challenges, we propose UR-BERT\footnote{Official implementation: https://github.com/sanghyang00/ur-bert}, a speech-aware pretrained text encoder for massively multilingual TTS covering 495 languages. We adopt Romanization as a language-agnostic textual interface in place of language-specific G2P systems, enabling scalable coverage beyond the limitations of existing G2P pipelines.
To further enhance phonetic modeling, we introduce a knowledge distillation objective based on speech token prediction. Specifically, a multilingual speech self-supervised model (S3M) serves as the teacher, and UR-BERT is trained to predict its output tokens, aligning textual representations with rich acoustic latent spaces. This alignment mitigates the phonetic abstraction introduced by Romanization and narrows the text-speech modality gap, achieving both scalability across languages and high phonetic fidelity.

In experiments, UR-BERT consistently outperforms prior BERT-style TTS encoders across a broad range of languages and evaluation metrics in both high- and low-resource settings, while supporting substantially more languages without compromising synthesis quality. 
Furthermore, the model maintains strong performance even with reduced amounts of pretraining data, underscoring the effectiveness of integrating Romanization with the proposed speech-aware pretraining strategy.

\begin{figure*}[t!]
    \centering
    \includegraphics[width=0.9\textwidth]{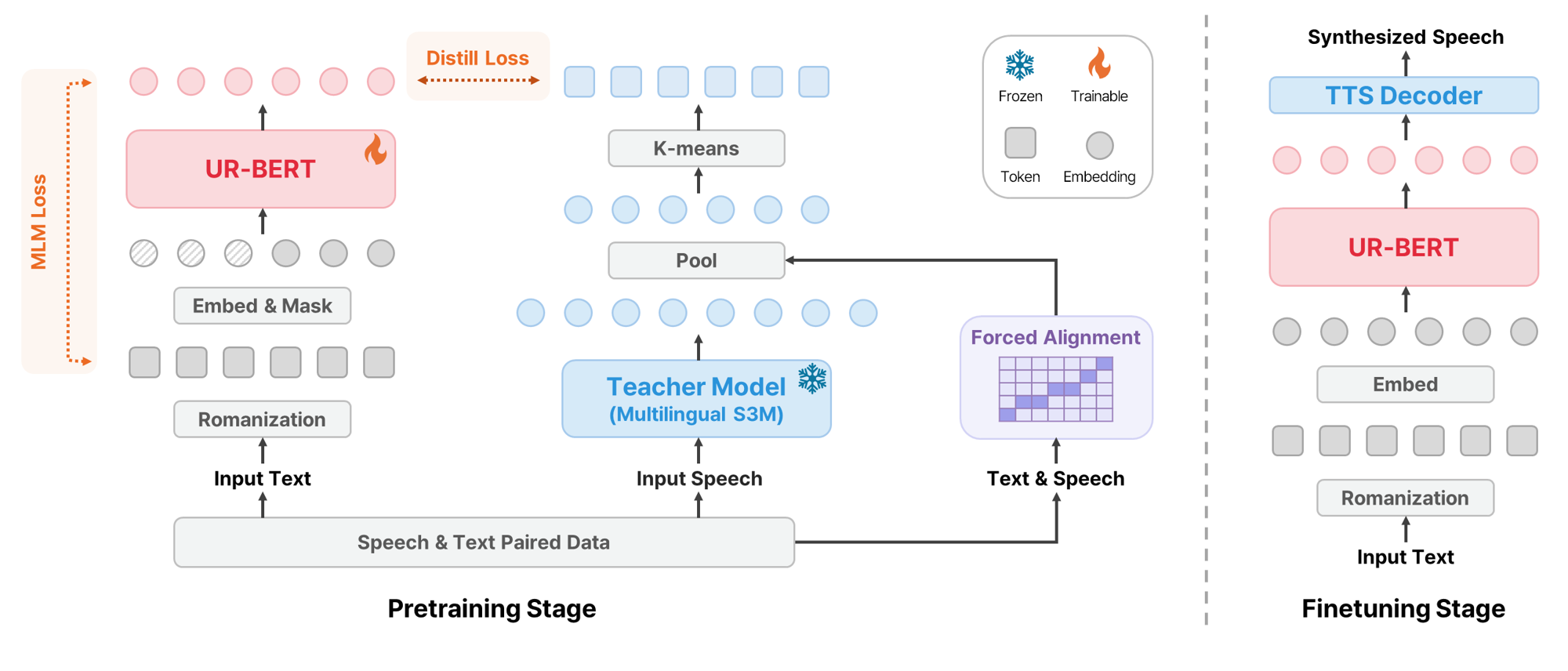}
    \vspace{-10pt}
    \caption{Overview of the UR-BERT showing pretraining and finetuning stage.}
    \label{fig:model}
    \vspace{-10pt}
\end{figure*}

\noindent Our contributions are summarized as follows:
\begin{itemize}
    \item We propose UR-BERT, a multilingual text encoder for TTS, pretrained on speech–text pairs covering 495 languages.
    \item We overcome the language coverage limitations of existing G2P pipelines by adopting Romanization as a unified orthographic interface for massively multilingual TTS.
    \item We introduce a novel speech token prediction-based pretraining strategy that aligns BERT-style text representations with acoustic information, enabling high-quality TTS synthesis.
\end{itemize}
\section{Related Work}
To extend monolingual text embeddings to multilingual TTS encoders, recent work has adopted BERT-style pretraining for text representations.
An early effort in this direction is multilingual PLBERT (m-PLBERT), introduced in the StyleTTS2~\cite{styletts2} framework.\footnote{https://huggingface.co/papercup-ai/multilingual-pl-bert} Following the original PL-BERT~\cite{plbert} design, m-PLBERT pretrains the text encoder on phoneme sequences from 15 languages, generated using Phonemizer~\cite{phonemizer}. However, its language coverage is limited to relatively high-resource languages, including English, Chinese, and several European languages.
Subsequently, XPhoneBERT~\cite{xphonebert} extended this paradigm by pretraining on phoneme sequences from 88 languages using CharsiuG2P~\cite{charsiug2p}. While it substantially increases language coverage, the pretraining data remain concentrated in European and Asian languages, with limited representation of many African and Indigenous American languages.

Scaling these approaches to a truly massive number of languages remains challenging due to their strong reliance on G2P systems. High-quality rule-based G2P modules are scarce, and even existing toolkits cover only a small fraction of the world’s languages. Moreover, zero-shot neural G2P alternatives often exhibit unstable performance, further limiting their applicability to previously unseen languages.

\section{Proposed Method}
\subsection{Architecture Overview}
The key distinctions of the proposed UR-BERT lie in its language scalability and training objectives. UR-BERT adopts Romanization as a unified text representation, enabling scalable modeling across diverse writing systems without reliance on G2P systems. It is pretrained on speech–text paired data spanning 495 languages, using a standard BERT-base architecture~\cite{bert} with a character-level tokenizer and 12 Transformer encoder layers~\cite{transformer}. In addition to the conventional masked language modeling (MLM) objective, UR-BERT incorporates speech token prediction (STP) as an auxiliary objective, injecting text-conditioned acoustic information during pretraining.
Figure~\ref{fig:model} illustrates the pretraining and fine-tuning pipeline of UR-BERT, with detailed design choices described in the following subsections.

\subsection{Romanization for Language Scalability}
We adopt Romanization to unify diverse orthographic systems into the Latin alphabet due to its superior scalability and token efficiency compared to phoneme-based approaches.
Phoneme-based methods rely on G2P systems, which require substantial linguistic expertise to design fine-grained, language-specific rules, thereby limiting practical coverage. As a result, existing G2P toolkits support only around 100 languages, such as 88 languages in CharsiuG2P~\cite{charsiug2p} and 127 languages in Phonemizer~\cite{phonemizer}.
In contrast, Romanization enables theoretically unbounded scalability by transliterating diverse writing systems into a shared Latin script, as exemplified by the Uroman toolkit~\cite{uroman}.
This advantage has been shown to scale to thousands of languages across multiple tasks, including TTS~\cite{xtts} and automatic speech recognition (ASR)~\cite{mms,lamaut}.

Conventional G2P systems convert graphemes into phonetic representations using the International Phonetic Alphabet (IPA)~\cite{ipa}. While IPA representations provide fine-grained phonetic detail, they require a large and diverse symbol inventory, often spanning thousands of symbols, which substantially increases vocabulary size and complicates tokenization. For example, some tokenization schemes treat prosodic markers, such as suprasegmentals and diacritics, as independent tokens despite their lack of standalone phonetic meaning, whereas others merge them with neighboring vowel or consonant tokens leading to inconsistent token granularity.
In contrast, Romanization transliterates non-Latin scripts into Latin characters, limiting the token inventory to approximately 30 alphabetic symbols and avoiding explicit prosodic markers. This compact token space simplifies tokenization and promotes more stable training. Moreover, prior work has shown that Romanization retains sufficient phonetic information for a wide range of speech-related tasks~\cite{xtts,mms,lamaut}, despite the reduced vocabulary size.

\subsection{Speech Token Injection for Phonetic Fidelity}
Despite the advantages of Romanization, capturing fine-grained phonetic distinctions remains challenging due to the limited token inventory compared to IPA, particularly when identical Romanized representation correspond to different pronunciations across languages.
To mitigate this acoustic ambiguity, we introduce a knowledge distillation that injects acoustic token information from a pretrained multilingual speech self-supervised model (S3M)~\cite{xlsr53,xlsr,mms,xeus,omnilingualasr} into UR-BERT during pretraining.

Unlike conventional TTS systems that require clean, curated speech data, our approach leverages large-scale ASR speech–text pairs by injecting speech-derived supervision into the text encoder through three steps: (1) extracting speech representations from S3M, (2) aligning them to character-level text using forced alignment, and (3) discretizing the aligned representations into speech tokens that serve as auxiliary training targets.
Through this process, ASR corpora are reframed as a scalable source of phonetic guidance, enabling TTS models to benefit from data previously unsuitable for speech synthesis.

\noindent\textbf{Speech Representation Extraction.}
We employ the omnilingual-ASR-W2V-300M model~\footnote{https://huggingface.co/facebook/omniASR-W2V-300M} as the teacher network and extract representations from its 16th layer. This design choice is motivated by prior findings that intermediate layers of multilingual S3Ms predominantly encode phonetic-level information rather than high-level semantic representations~\cite{layerwise1,layerwise2}.

\begin{figure}[t!]
    \centering
    \includegraphics[width=0.9\columnwidth]{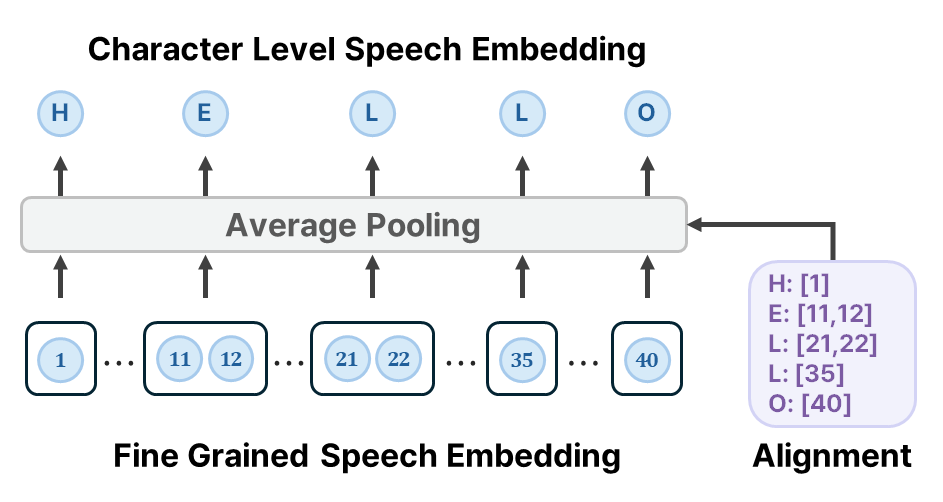}
    \vspace{-10pt}
    \caption{Illustration of the CTC-based speech-text alignment.}
    \label{fig:alignment}
    \vspace{-15pt}
\end{figure}
\noindent\textbf{CTC-Based Speech-Text Alignment.}
A key challenge in speech-text alignment arises from the mismatch in sequence length between speech and text, as acoustic feature sequences are typically much longer than their textual counterparts.
To obtain character-level acoustic representations, we apply CTC-based forced alignment using MMS-FA~\cite{mms}, followed by average pooling over the aligned frames for each character. 
The overall alignment procedure is illustrated in Figure~\ref{fig:alignment}.

\noindent\textbf{Discrete Token Assignment.}
To discretize continuous character-level acoustic representations, we perform $k$-means clustering over the pretraining corpus to construct a finite codebook. Each character-level acoustic representation is assigned to its nearest cluster centroid, yielding a discrete speech token for every Romanized character. These tokens are used as supervision for the STP objective during pretraining, enabling UR-BERT to infer acoustic information directly from text input.
We set the codebook size to 257, where index 0 represents a mute token, and indices 1-256 correspond to acoustic tokens. Larger codebooks were not considered, as excessive capacity tends to encode speaker-dependent or paralinguistic variations rather than phonetic content~\cite{selm,diffkmeans}, and may destabilize training due to the mismatch with the compact text vocabulary.
This design choice is further motivated by phonological theory, which models speech sound inventories as economical combinations of a limited number of binary features~\cite{tokenlimit}, striking a balance between representational capacity and phonetic abstraction.

\section{Experiments}
\subsection{Pretraining}
\label{sec:setup}
We construct the pretraining corpus by combining three ASR datasets: FLEURS~\cite{fleurs}, which spans 102 read-speech languages; Common Voice~\cite{commonvoice}, a crowdsourced dataset covering 131 languages; and the Omnilingual ASR corpus~\cite{omnilingualasr}, which includes 348 low-resource languages. The resulting pretraining corpus contains approximately 13K hours of speech across 495 languages, comprising 8M sentences as summarized in Table~\ref{tab:model_config}. Pretraining is conducted for 150K steps with a batch size of 1024 using gradient accumulation. We employ the AdamW~\cite{adamw} optimizer with a tri-stage learning rate schedule~\cite{data2vec,wav2vec2,wavlm}, using warm-up, peak, and decay ratios of 0.1, 0.5, and 0.4 with a peak learning rate of $1\text{e-}4$.

\subsection{TTS Finetuning}
We conduct downstream TTS experiments on 11 languages spanning both high- and low-resource settings, with all datasets resampled to 22,050 Hz.
The high-resource group includes English~\cite{ljspeech}, German~\cite{thorsten}, and Mandarin Chinese~\cite{aishell3}, each with 20 hours of training data. 
The low-resource group consists of eight Asian and African languages: Javanese, Sundanese, Khmer, Nepali~\cite{lowresourcedb1}, Sinhala~\cite{lowresourcedb2}, Afrikaans, Setswana, and Xhosa~\cite{lowresourcedb3}.
Specifically, we used 5 hours of training data for Javanese and Sundanese, 3 hours for Khmer, 2 hours for Afrikaans, Nepali, Setswana, and Xhosa, and 1 hour for Sinhala, reflecting differences in available data per language. 

For TTS modeling, we adopt VITS~\cite{vits} as the backbone architecture and compare its original text encoder with existing BERT-style encoders, including m-PLBERT and XPhoneBERT, as well as the proposed UR-BERT. 
Low-resource models are trained for 100K steps, while high-resource models are trained for 300K steps, both with a batch size of 32, following the training protocols of MMS-TTS and XPhoneBERT, respectively. 
The optimization settings largely follow the pretraining configuration, except that the text encoder is frozen for the first 25\% of training steps and the warm-up schedule is omitted to stabilize the monotonic alignment search module. 

\begin{table}[!t]
\centering
\caption{Comparison of baselines and UR-BERT. Dataset amounts are measured with sentences and hours, respectively.}
\vspace{-7pt}
\begin{threeparttable}
\resizebox{0.95\columnwidth}{!}{%
\begin{tabular}{l|cc|cc} \toprule
\multirow{2}{*}{\textbf{Model}} & \multicolumn{2}{c|}{\textbf{Languages}} & \multicolumn{2}{c}{\textbf{Datasets}} \\ \cmidrule{2-3} \cmidrule{4-5}
 & \textbf{\# Train} & \textbf{\# Support} & \textbf{\# Text} & \textbf{\# Speech} \\ \midrule\midrule
m-PLBERT   & 15  & 127  & 150K & - \\
XPhoneBERT & 88  & 99   & 330M & - \\ \midrule
\textbf{UR-BERT}    & \textbf{495} & \textbf{1162}\tnote{$\star$} & \textbf{8M} & \textbf{13K} \\ \bottomrule
\end{tabular}
}

\begin{tablenotes}[flushleft]
\scriptsize
\item[$\star$] While the romanization toolkit (Uroman) is theoretically language-agnostic, we report the maximum number of languages empirically validated in prior studies.
\end{tablenotes}
\end{threeparttable}

\vspace{-15pt}
\label{tab:model_config}
\end{table}
\begin{table*}[!t]
\centering
\caption{Performance on high-resource languages. MPB, XPB, and URB denote m-PLBERT, XPhoneBERT, and UR-BERT, respectively. $\Delta$CER is reported in percentage points, and F0 denotes log-$F0$ RMSE. Best results are bolded, and second-best are underlined.}
\vspace{-7pt}
\resizebox{0.98\textwidth}{!}{%
\begin{tabular}{l|ccccc|ccccc|ccccc} \toprule
\multirow{2}{*}{\textbf{Model}} & \multicolumn{5}{c|}{\textbf{English (EN)}} & \multicolumn{5}{c|}{\textbf{German (DE)}} & \multicolumn{5}{c}{\textbf{Mandarin Chinese (ZH)}} \\
                       & \textbf{MOS} $\uparrow$ & $\Delta$\textbf{UTM} $\downarrow$ & $\Delta$\textbf{CER} $\downarrow$ & \textbf{MCD} $\downarrow$ & \textbf{F0} $\downarrow$ & \textbf{MOS} $\uparrow$ & $\Delta$\textbf{UTM} $\downarrow$ & $\Delta$\textbf{CER} $\downarrow$ & \textbf{MCD} $\downarrow$ & \textbf{F0} $\downarrow$ & \textbf{MOS} $\uparrow$ & $\Delta$\textbf{UTM} $\downarrow$ & $\Delta$\textbf{CER} $\downarrow$ & \textbf{MCD} $\downarrow$ & \textbf{F0} $\downarrow$ \\ \midrule\midrule
GT                     & 4.61 & - & - & - & - & 4.02 & - & - & - & - & 4.35 & - & - & - & - \\
VITS                   & 3.78 & 0.29 & 6.15 & 5.71 & 0.162 & 3.45 & 0.53 & 6.37 & 5.36 & 0.188 & \underline{3.65} & 0.46 & 29.28 & 5.17 & 0.102 \\
+MPB                   & 1.83 & 0.58 & 66.50 & 8.34 & 0.163 & 2.65 & 0.57 & 67.78 & 6.93 & 0.203 & 2.88 & \underline{0.35} & 67.45 & 5.90 & 0.105 \\
+XPB                   & \underline{4.11} & \underline{0.23} & \underline{4.79} & \textbf{5.17} & \underline{0.152} & \underline{3.53} & \underline{0.53} & \underline{5.85} & \underline{4.77} & \underline{0.171} & 3.49 & \textbf{0.30} & \underline{25.98} & \underline{5.16} & \textbf{0.097} \\ \midrule
\textbf{+URB}                   & \textbf{4.35} & \textbf{0.12} & \textbf{3.78} & \underline{5.23} & \textbf{0.149} & \textbf{3.78} & \textbf{0.33} & \textbf{3.07} & \textbf{4.66} & \textbf{0.170}  & \textbf{3.88} & 0.36 & \textbf{21.83} & \textbf{4.95} & \underline{0.098}\\ \bottomrule
\end{tabular}%
}
\label{tab:high_resource}
\vspace{-8pt}
\end{table*}

\begin{table}[!t]
\centering
\caption{Performance on low-resource languages. We denote the dataset size for each language, and best results are bolded.}
\vspace{-7pt}
\resizebox{0.95\columnwidth}{!}{%
\begin{tabular}{l|l|ccccc} \toprule
\textbf{Language} & \textbf{Model} & \textbf{MOS} $\uparrow$ & $\Delta$\textbf{UTM} $\downarrow$ & $\Delta$\textbf{CER} $\downarrow$ & \textbf{MCD} $\downarrow$ & \textbf{F0} $\downarrow$ \\ \midrule\midrule
\rowcolor[gray]{0.9} \multicolumn{7}{c}{\textit{Group 1: Seen languages supported by XPhoneBERT and UR-BERT}} \\ \midrule
\multirow{4}{*}{\shortstack[l]{Afrikaans\\(AF, 2h)}} & GT & 4.21 & - & - & - & - \\
 & VITS & 2.80 & 0.59 & 26.03 & 6.41 & 0.127 \\
 & +XPB & 2.85 & 0.38 & 18.54 & 6.25 & 0.122 \\ \cmidrule{2-7}
 & \textbf{+URB} & \textbf{3.34} & \textbf{0.37} & \textbf{15.82} & \textbf{6.09} & \textbf{0.121} \\ \midrule
\multirow{4}{*}{\shortstack[l]{Khmer\\(KM, 3h)}} & GT & 3.58 & - & - & - & - \\
 & VITS & 2.96 & 0.51 & 15.51 & 6.17 & 0.117 \\
 & +XPB & 2.98 & \textbf{0.50} & 12.40 & 5.72 & 0.113 \\ \cmidrule{2-7}
 & \textbf{+URB} & \textbf{3.21} & 0.52 & \textbf{6.88} & \textbf{5.59} & \textbf{0.112} \\ \midrule\midrule
\rowcolor[gray]{0.9} \multicolumn{7}{c}{\textit{Group 2: Seen languages exclusively supported by UR-BERT}} \\ \midrule
\multirow{3}{*}{\shortstack[l]{Javanese\\(JV, 5h)}} & GT & 3.90 & - & - & - & - \\
 & VITS & 2.80 & 0.61 & \textbf{23.50} & \textbf{6.42} & 0.107 \\ \cmidrule{2-7}
 & \textbf{+URB} & \textbf{3.05} & \textbf{0.52} & 28.03 & 6.64 & \textbf{0.105} \\ \midrule
\multirow{3}{*}{\shortstack[l]{Nepali\\(NP, 2h)}} & GT & 4.35 & - & - & - & - \\
 & VITS & 3.33 & 0.47 & 13.55 & 7.10 & 0.071 \\ \cmidrule{2-7}
 & \textbf{+URB} & \textbf{3.66} & \textbf{0.46} & \textbf{6.71} & \textbf{6.76} & \textbf{0.068} \\ \midrule
\multirow{3}{*}{\shortstack[l]{Setswana\\(TN, 2h)}} & GT & 3.92 & - & - & - & - \\
 & VITS & 2.51 & 0.78 & 32.27 & 5.38 & 0.118 \\ \cmidrule{2-7}
 & \textbf{+URB} & \textbf{2.92} & \textbf{0.77} & \textbf{25.50} & \textbf{5.13} & \textbf{0.113} \\ \midrule
\multirow{3}{*}{\shortstack[l]{Xhosa\\(XH, 2h)}} & GT & 4.20 & - & - & - & - \\
 & VITS & 3.05 & 0.58 & 17.70 & 6.96 & 0.112 \\ \cmidrule{2-7}
 & \textbf{+URB} & \textbf{3.48} & \textbf{0.54} & \textbf{10.74} & \textbf{6.13} & \textbf{0.106} \\ \midrule
\multirow{3}{*}{\shortstack[l]{Sinhala\\(SI, 1h)}} & GT & 4.11 & - & - & - & - \\
 & VITS & 3.49 & 0.48 & 13.09 & 5.31 & 0.095 \\ \cmidrule{2-7}
 & \textbf{+URB} & \textbf{3.82} & \textbf{0.39} & \textbf{9.49} & \textbf{4.75} & \textbf{0.091} \\ \midrule\midrule
\rowcolor[gray]{0.9} \multicolumn{7}{c}{\textit{Group 3: Unseen language for UR-BERT (Zero-shot)}} \\ \midrule
\multirow{3}{*}{\shortstack[l]{Sundanese\\(SU, 5h)}} & GT & 4.25 & - & - & - & - \\
 & VITS & 3.15 & 0.59 & 14.86 & \textbf{4.98} & 0.078 \\ \cmidrule{2-7}
 & \textbf{+URB} & \textbf{3.43} & \textbf{0.47} & \textbf{13.46} & 5.08 & \textbf{0.077} \\ \bottomrule
\end{tabular}%
}
\label{tab:low_resource_single}
\vspace{-10pt}
\end{table}
\subsection{Performance Metrics}
We evaluate UR-BERT using one subjective and four objective metrics to provide a comprehensive assessment.
For subjective evaluation, we conduct a standard mean opinion score (MOS) test on a 1-5 scale with phonetic guidance. Ratings are collected on 520 samples from 44 participants with diverse regional backgrounds. 
For objective quality assessment, we employ the UTokyo MOS Prediction System (UTMOS)~\cite{voicemos, utmos}. To mitigate potential cross-lingual bias, we report relative degradation ($\Delta$UTM) with respect to ground-truth (GT) samples.
Intelligibility is evaluated using character error rate (CER), computed from transcriptions generated by Omnilingual-ASR-CTC-1B~\footnote{https://huggingface.co/facebook/omniASR-CTC-1B}, and is similarly reported as relative degradation ($\Delta$CER) against GT speech. 
To quantify spectral and prosodic differences between synthesized and GT speech, we additionally report mel-cepstral distance (MCD) and log-$F0$ root mean squared error (Log-$F0$ RMSE).
\section{Results}
\subsection{Performance on High-Resource Languages}
Table~\ref{tab:high_resource} presents TTS evaluation results on high-resource languages. While VITS achieves strong baseline performance in these settings, incorporating UR-BERT consistently improves both subjective and objective metrics across all evaluated languages.
In contrast, m-PLBERT exhibits notable performance degradation when integrated with VITS, often producing fluent but phonetically inaccurate speech, resulting in elevated CER.
XPhoneBERT achieves competitive performance as a result of large-scale multilingual pretraining; however, it demonstrates lower naturalness and intelligibility compared to UR-BERT.

The performance gains of UR-BERT are particularly notable given that it is trained on only 2.5\% of the data used by XPhoneBERT (8M vs. 330M sentences). We attribute this efficiency to the proposed framework, which combines a compact token space enabled by Romanization with an STP objective. The reduced vocabulary facilitates data-efficient learning, while speech-aware pretraining enhances text–speech alignment and enriches phonetic representations. 

\newcommand{\cmark}{\ding{51}}
\newcommand{\xmark}{\ding{55}}
\begin{table}[!t]
\centering
\caption{Ablation study on STP; We report the MOS of the UR-BERT with and without the STP. Bold indicates the higher value.}
\vspace{-7pt}
\resizebox{0.99\columnwidth}{!}{
\begin{tabular}{c|ccc|cccccccc} \toprule
\multirow{2}{*}{\textbf{STP}} & \multicolumn{3}{c|}{\textbf{High-resource}} & \multicolumn{7}{c}{\textbf{Low-resource}} \\
 & \textbf{EN} & \textbf{DE} & \textbf{ZH} & \textbf{AF} & \textbf{KM} & \textbf{JV} & \textbf{NP} & \textbf{TN} & \textbf{XH} & \textbf{SI} & \textbf{SU} \\ \midrule\midrule
\cmark & \textbf{4.35} & \textbf{3.78} & \textbf{3.88} & \textbf{3.34} & 3.21 & \textbf{3.05} & \textbf{3.66} & \textbf{2.92} & \textbf{3.48} & \textbf{3.82} & \textbf{3.43} \\
\xmark & 4.00 & 3.64 & 3.75 & 3.04 & \textbf{3.23} & 2.65 & \textbf{3.66} & 2.70 & 3.34 & 3.52 & 3.41 \\ \bottomrule
\end{tabular}
}
\label{tab:ablation}
\vspace{-15pt}
\end{table}
\subsection{Performance on Low-Resource Languages}
Table~\ref{tab:low_resource_single} presents TTS performance on low-resource languages. Group 1 includes languages supported by both XPhoneBERT and UR-BERT, whereas Group 2 consists of languages supported only by UR-BERT. Phoneme-based baselines often fail to support many low-resource languages due to the absence of reliable G2P systems. In contrast, UR-BERT naturally extends to these languages through Romanization, demonstrating strong scalability.
Across both groups, UR-BERT consistently achieves the strongest overall performance, obtaining the highest MOS for every language and outperforming baselines on most objective metrics. The gains are particularly pronounced in terms of naturalness and intelligibility.

To further evaluate cross-lingual generalization of UR-BERT, we conduct additional experiments on Sundanese (Group 3), which is excluded from the pretraining corpus. Despite this zero-shot setting, UR-BERT consistently outperforms baseline models, demonstrating robust generalization to languages unseen during pretraining.

\subsection{Effectiveness of Speech Token Prediction Objective}
Table~\ref{tab:ablation} shows that removing the STP objective consistently degrades performance across nearly all languages, with noticeable MOS drops in both settings. The effect is particularly pronounced for high-resource languages, while remaining evident in low-resource and zero-shot scenarios.
These results indicate that STP is critical for enriching phonetic representations and stabilizing text–speech alignment, effectively compensating for the phonetic abstraction inherent in Romanization.
\section{Conclusion}
In this paper, we propose UR-BERT, a multilingual and multimodal pretrained text encoder for text-to-speech applications. By adopting Romanization as a unified text representation, UR-BERT overcomes the language coverage limitations of conventional G2P pipelines and enables scalable pretraining across 495 languages. To further enhance phonetic fidelity, we introduce a speech-token prediction objective that injects acoustic knowledge into text representations and strengthens text–speech modality alignment. Experimental results demonstrate that UR-BERT consistently outperforms prior approaches across languages with varying resource availability, while exhibiting strong cross-lingual generalization. We envision UR-BERT as a foundational building block toward truly universal, massively multilingual TTS systems, enabling scalable and inclusive speech synthesis across the world’s languages.
\section{Generative AI Use Disclosure}
All co-authors attest that generative AI tools were employed exclusively to refine human-authored text and to support LaTeX formatting of the manuscript, including tables and figures. No generative AI tools were used in the development of research ideas, analytical procedures, or the creation of any substantive scientific content. We further reaffirm our commitment to the responsible and ethical use of generative AI in accordance with established research ethics.
\section{Acknowledgement}
This work was supported by the National Research Foundation of Korea (NRF) grant funded by the Korea government Ministry of Science and ICT (MSIT) (RS-2026-25468664). 

\bibliographystyle{IEEEtran}
\bibliography{mybib}

\clearpage
\appendix
\twocolumn[{
    \part*{Appendix}
    \begin{center}
        \centering
        \begin{minipage}{0.48\textwidth}
            \centering
            \includegraphics[width=\textwidth]{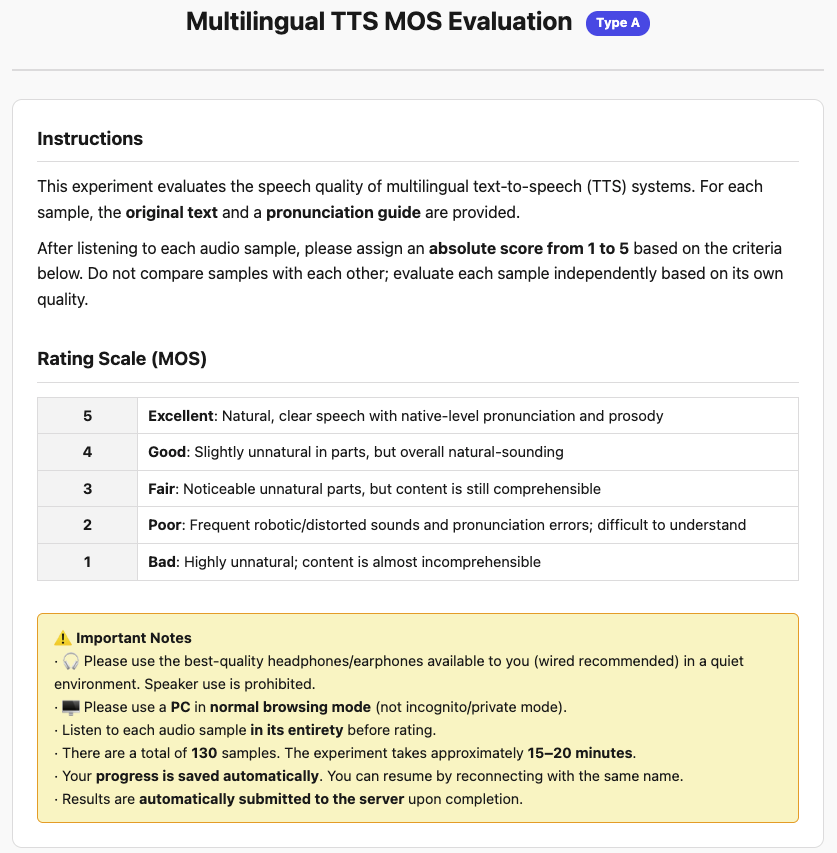}
            \small (a) Instructions provided to the participants. 
            \label{appendix:mos_instruction}
        \end{minipage}
        \hfill 
        \begin{minipage}{0.48\textwidth}
            \centering
            \includegraphics[width=\textwidth]{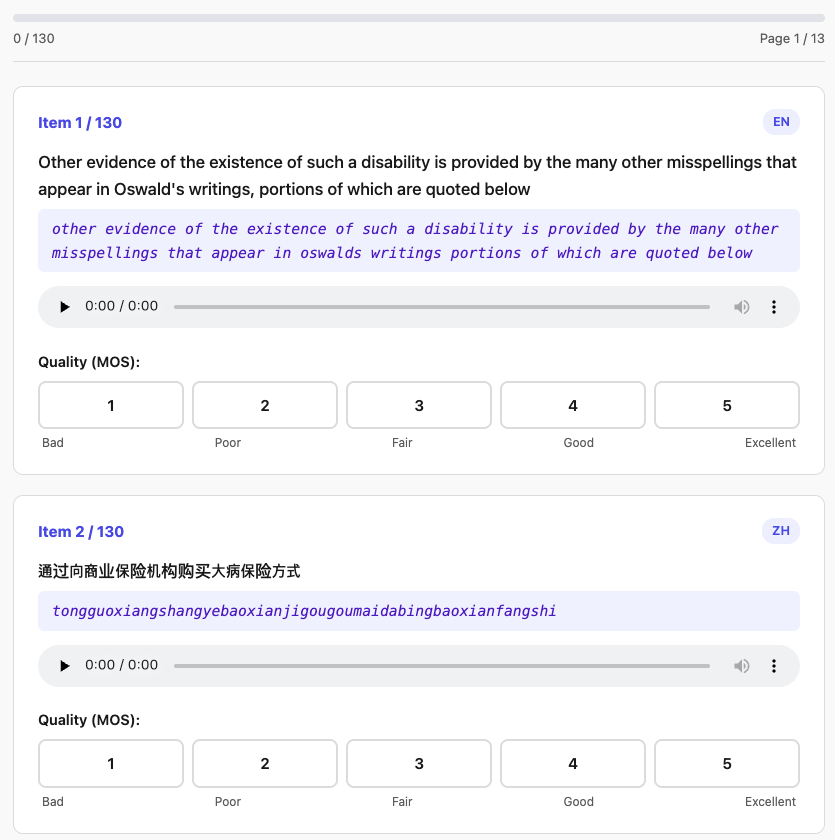}
            \small (b) Interface of the MOS evaluation platform. 
            \label{appendix:mos_protocol}
        \end{minipage}
        \vspace{-2pt}
        \captionof{figure}{MOS evaluation setup: (a) detailed instructions for quality assessment, and (b) a snapshot of the evaluation interface.}
        \label{appendix:mos}
        \vspace{2pt}
    \end{center}
}]

\section{Details on the Pretraining Dataset}
\label{appendix:config}

\noindent\textbf{Preprocessing.}
We removed samples whose transcriptions contained digits or parenthetical expressions. Digits may correspond to different pronunciations across languages, leading to token–pronunciation mismatches, while parenthetical content is inconsistently realized in speech. 
The Omnilingual ASR corpus generally contains longer utterances, as its collection protocol was designed to elicit natural, prompt-based responses. To reduce computational overhead and excessive padding during training, we segmented the Omnilingual ASR samples into chunks of up to 30 seconds using MMS-FA, aligning their duration distribution with that of FLEURS and CommonVoice.

\noindent\textbf{Configuration.}
As described in Section~\ref{sec:setup}, we combined three ASR-oriented speech–text paired datasets to construct a large-scale pretraining corpus for UR-BERT. Language-specific details, including language names, ISO-639-3 codes, and dataset sizes, are provided in Table~\ref{tab:appendix_config}, listed in alphabetical order.

\section{Details on MOS Evaluation Protocol}
\label{appendix:mos}
\noindent\textbf{Participants.}
We recruited 44 participants through community outreach. Given the multilingual scope of our evaluation, which spans 11 languages across diverse geographic regions (e.g., the United States, the United Kingdom, Germany, China, as well as regions in Africa and Asia), we selected participants with demonstrated familiarity with at least a subset of the evaluated languages. All participants were fluent in English, and several had lived in or had prior exposure to other language regions (e.g., Germany or Indonesia), ensuring reliable judgments.

\noindent\textbf{Evaluation Procedure.}
To facilitate faithful evaluation, we provided both grapheme transcriptions and Romanized transcriptions for each utterance as phonetic guidance, allowing participants to assess whether the synthesized speech aligned with the intended text. Language information was also displayed for each sample. Participants were instructed to evaluate in a quiet environment using high-quality personal audio equipment (e.g., headphones). The use of loudspeakers was prohibited to avoid variability in acoustic perception. The detailed survey interface is illustrated in Figure~\ref{appendix:mos}.

\begin{center}
    \centering
    \vspace{-7pt}
    \resizebox{0.9\columnwidth}{!}{%
    \begin{tabular}{l|c|c} 
    \toprule
    \textbf{Model} & \textbf{Language Coverage} & \textbf{\# Lang.} \\ 
    \midrule\midrule
    VITS       & \begin{tabular}[c]{@{}c@{}}EN, ZH, DE, JV, SU, KM,\\ NP, AF, TN, XH, SI\end{tabular} & 11 \\ \midrule
    $+$m-PLBERT   & EN, ZH, DE                                                   & 3  \\ \midrule
    $+$XPhoneBERT & EN, ZH, DE, KM, AF                                           & 5  \\ \midrule
    
    \textbf{$+$UR-BERT} & \multirow{3}{*}{\begin{tabular}[c]{@{}c@{}}EN, ZH, DE, JV, SU, KM,\\ NP, AF, TN, XH, SI\end{tabular}} & \multirow{3}{*}{\begin{tabular}[c]{@{}c@{}}11 each\\ (Total 22)\end{tabular}} \\ 
    \quad $\bullet$ with ATP  &  & \\
    \quad $\bullet$ without ATP &  & \\ 
    \bottomrule
    \end{tabular}%
    }
    \vspace{2pt}
    \captionof{table}{Details of the TTS models; 41 models in total.}
    \label{tab:appendix_model}
    \vspace{-15pt}
\end{center}

\noindent\textbf{Stimuli and Assignment.}
We randomly sampled 10 utterances per configuration. Sample identities were shared across models within each language to enable fair comparison (e.g., VITS, m-PLBERT, XPhoneBERT, and UR-BERT variants for English used identical text prompts). As shown in Table~\ref{tab:appendix_model}, this resulted in 520 evaluated samples from 110 ground truth and 410 generated samples.
To mitigate listener fatigue and maintain consistent scoring criteria, the samples were divided into four non-overlapping groups of 130 utterances each (Groups A, B, C, and D). Participants were evenly assigned to groups (11 per group), ensuring that no evaluator assessed overlapping samples.

\begingroup
\onecolumn
\centering
\tiny
\setlength{\tabcolsep}{2pt}

\tablehead{
    \toprule
    \textbf{Language} & \textbf{ISO Code} & \textbf{Duration} & \textbf{Sentence} & 
    \textbf{Language} & \textbf{ISO Code} & \textbf{Duration} & \textbf{Sentence} & 
    \textbf{Language} & \textbf{ISO Code} & \textbf{Duration} & \textbf{Sentence} \\
    \midrule\midrule
}

\tabletail{\midrule \multicolumn{12}{r}{\scriptsize Continued on next page...} \\ \bottomrule}

\tablelasttail{\bottomrule}

\begin{supertabular}{lccc|lccc|lccc}
Abadi                          & kbt      & 5.91     & 1828     & Abkhazian                               & abk      & 31.86    & 21037    & Abron                            & abr      & 5.10     & 1908     \\
Abua                           & abn      & 6.34     & 2098     & Afade                                   & aal      & 6.88     & 2073     & Afrikaans                        & afr      & 2.41     & 939      \\
Agwagwune                      & yay      & 4.62     & 1753     & Akan                                    & aka      & 0.22     & 205      & Akebu                            & keu      & 6.72     & 2389     \\
Alago                          & ala      & 6.83     & 2138     & Algerian Arabic                         & arq      & 9.66     & 2081     & Ambonese Malay                   & abs      & 6.49     & 2216     \\
Amharic                        & amh      & 9.13     & 2955     & Anaang                                  & anw      & 6.87     & 2696     & Angika                           & anp      & 5.07     & 1404     \\
Antankarana Malagasy           & xmv      & 12.75    & 2589     & Arabic, Algerian Saharan                & aao      & 1.96     & 449      & Arabic, Dhofari                  & adf      & 0.31     & 66       \\
Arabic, Judeo-Moroccan         & aju      & 7.14     & 1235     & Arbëreshë Albanian                      & aae      & 9.75     & 3788     & Armenian                         & hye      & 21.28    & 11668    \\
Ashe                           & ahs      & 7.03     & 2099     & Askopan                                 & eiv      & 5.64     & 1548     & Assamese                         & asm      & 9.04     & 3029     \\
Asturian                       & ast      & 6.16     & 2376     & Awak                                    & awo      & 6.06     & 2010     & Ayacucho Quechua                 & quy      & 0.04     & 26       \\
Azerbaijani                    & aze      & 7.11     & 2166     & Bacama                                  & bcy      & 5.57     & 1801     & Bade                             & bde      & 6.07     & 2127     \\
Bago-Kusuntu                   & bqg      & 6.55     & 2276     & Baharna Arabic                          & abv      & 10.26    & 1734     & Balanta-Ganja                    & bjt      & 6.37     & 2276     \\
Bangwinji                      & bsj      & 6.40     & 2287     & Banjar                                  & bjn      & 6.72     & 2561     & Bara Malagasy                    & bhr      & 11.89    & 2698     \\
Barok                          & bjk      & 6.21     & 1597     & Basa (Cameroon)                         & bas      & 2.16     & 2109     & Basa (Nigeria)                   & bzw      & 6.56     & 2443     \\
Bashkir                        & bak      & 142.96   & 119000   & Basque                                  & eus      & 205.87   & 130034   & Batak Mandailing                 & btm      & 6.67     & 2329     \\
Bayot                          & bda      & 6.01     & 1461     & Belarusian                              & bel      & 483.05   & 349582   & Bengali                          & ben      & 41.95    & 23777    \\
Betawi                         & bew      & 6.76     & 2482     & Bhili                                   & bhb      & 6.75     & 1922     & Bhojpuri                         & bho      & 5.38     & 1335     \\
Bilur                          & bxf      & 6.36     & 1864     & Bima                                    & bhp      & 6.47     & 1805     & Bodo (India)                     & brx      & 6.52     & 1957     \\
Boghom                         & bux      & 5.32     & 1870     & Bokyi                                   & bky      & 5.34     & 1761     & Bomu                             & bmq      & 10.24    & 2649     \\
Bondei                         & bou      & 6.19     & 1948     & Borgu Fulfulde                          & fue      & 9.35     & 2218     & Bosnian                          & bos      & 7.56     & 2427     \\
Brahui                         & brh      & 5.59     & 1879     & Braj                                    & bra      & 6.32     & 1975     & Breton                           & bre      & 3.43     & 3509     \\
Buduma                         & bdm      & 6.34     & 2018     & Buginese                                & bug      & 7.23     & 1739     & Bukharic                         & bhh      & 7.44     & 1587     \\
Bulgarian                      & bul      & 14.18    & 7244     & Bundeli                                 & bns      & 5.55     & 1655     & Bura-Pabir                       & bwr      & 7.07     & 2449     \\
Burak                          & bys      & 6.46     & 2038     & Burmese                                 & mya      & 9.09     & 2353     & Cacaloxtepec Mixtec              & miu      & 5.99     & 1617     \\
Cakfem-Mushere                 & cky      & 5.51     & 1931     & Campidanese Sardinian                   & sro      & 6.36     & 2250     & Catalan                          & cat      & 1809.27  & 1203445  \\
Cebuano                        & ceb      & 9.25     & 2563     & Cen                                     & cen      & 6.15     & 2268     & Central Kurdish                  & ckb      & 16.95    & 10243    \\
Central Nahuatl                & nhn      & 5.02     & 1087     & Central Pame                            & pbs      & 6.38     & 1988     & Central Pashto                   & pst      & 19.77    & 7735     \\
Central-Eastern Niger Fulfulde & fuq      & 3.75     & 1029     & Chadian Arabic                          & shu      & 2.29     & 319      & Chichicapan Zapotec              & zpv      & 5.62     & 2445     \\
Chiga                          & cgg      & 7.47     & 2387     & Chimalapa Zoque                         & zoh      & 5.89     & 1463     & Chimborazo Highland Quichua      & qug      & 5.95     & 1731     \\
Chitwania Tharu                & the      & 5.42     & 1882     & Chuvash                                 & chv      & 1.99     & 1455     & Cibak                            & ckl      & 6.24     & 2096     \\
Coastal Konjo                  & kjc      & 6.19     & 1863     & Croatian                                & hrv      & 8.78     & 2684     & Cross River Mbembe               & mfn      & 5.90     & 2053     \\
Cuyamecalco Mixtec             & xtu      & 6.13     & 1621     & Czech                                   & ces      & 35.18    & 23927    & Dadiya                           & dbd      & 5.83     & 2021     \\
Danish                         & dan      & 9.91     & 5515     & Dazaga                                  & dzg      & 5.71     & 2179     & Deccan                           & dcc      & 7.07     & 2668     \\
Degema                         & deg      & 7.03     & 2162     & Dera (Nigeria)                          & kna      & 7.71     & 2869     & Dghwede                          & dgh      & 5.65     & 1930     \\
Dhivehi                        & div      & 3.81     & 2654     & Dijim-Bwilim                            & cfa      & 6.78     & 2135     & Dotyali                          & dty      & 7.41     & 2688     \\
Dutch                          & nld      & 59.96    & 45746    & Dyula                                   & dyu      & 0.15     & 88       & D˜uya                            & ldb      & 7.74     & 2199     \\
Eastern Bolivian Guaraní       & gui      & 22.35    & 3867     & Eastern Egyptian Bedawi Arabic          & avl      & 1.86     & 311      & Eastern Krahn                    & kqo      & 5.42     & 2409     \\
Eastern Mari                   & mhr      & 234.73   & 185245   & Eastern Yiddish                         & ydd      & 12.95    & 2713     & Eggon                            & ego      & 6.26     & 1962     \\
Egyptian Arabic                & arz      & 15.34    & 3089     & Ejagham                                 & etu      & 6.35     & 2354     & Eleme                            & elm      & 7.13     & 2060     \\
Eloyi                          & afo      & 7.19     & 2167     & Embu                                    & ebu      & 5.46     & 1694     & English                          & eng      & 1784.77  & 1129098  \\
Erzya                          & myv      & 1.97     & 1241     & Esan                                    & ish      & 5.89     & 1334     & Esperanto                        & epo      & 246.13   & 142968   \\
Extremaduran                   & ext      & 13.07    & 4382     & Fanti                                   & fat      & 11.69    & 2951     & Farefare                         & gur      & 7.14     & 2204     \\
Filipino                       & fil      & 5.89     & 1488     & Filomena Mata-Coahuitlán Totonac        & tlp      & 7.33     & 1995     & Finnish                          & fin      & 8.92     & 3961     \\
Fipa                           & fip      & 7.00     & 2053     & French                                  & fra      & 858.73   & 595563   & Fulah                            & ful      & 9.83     & 2486     \\
Fulfulde, Bagirmi              & fui      & 14.55    & 3622     & Galician                                & glg      & 100.05   & 70715    & Gambian Wolof                    & wof      & 6.50     & 2416     \\
Ganda                          & lug      & 119.34   & 68780    & Garhwali                                & gbm      & 14.80    & 3086     & Gbagyi                           & gbr      & 8.16     & 2723     \\
Gbari                          & gby      & 8.47     & 2619     & Geji                                    & gyz      & 6.76     & 2467     & Georgian                         & kat      & 96.35    & 63703    \\
German                         & deu      & 969.37   & 610065   & Geser-Gorom                             & ges      & 6.15     & 1463     & Gheg Albanian                    & aln      & 3.60     & 591      \\
Glavda                         & glw      & 6.90     & 2298     & Goan Konkani                            & gom      & 5.16     & 1103     & Goemai                           & ank      & 6.49     & 2396     \\
Gola                           & gol      & 5.29     & 1905     & Guarani                                 & grn      & 1.15     & 1049     & Guduf-Gava                       & gdf      & 8.03     & 2302     \\
Guerrero Amuzgo                & amu      & 6.74     & 2058     & Gujarati                                & guj      & 6.61     & 2392     & Gulf Arabic                      & afb      & 18.76    & 2866     \\
Gusii                          & guz      & 5.64     & 2320     & Gusilay                                 & gsl      & 5.90     & 1617     & Gweno                            & gwe      & 6.24     & 2172     \\
Güilá Zapotec                  & ztu      & 5.64     & 1999     & Hahon                                   & hah      & 6.14     & 1825     & Haitian                          & hat      & 0.02     & 11       \\
Hakha Chin                     & cnh      & 0.65     & 817      & Hakö                                    & hao      & 6.63     & 2039     & Halia                            & hla      & 6.21     & 1919     \\
Haroti                         & hoj      & 8.46     & 2060     & Hausa                                   & hau      & 11.86    & 4307     & Hawaiian                         & haw      & 6.16     & 1027     \\
Hebrew                         & heb      & 8.21     & 3518     & Herero                                  & her      & 6.02     & 2151     & Highland Konjo                   & kjk      & 6.26     & 2079     \\
Hijazi Arabic                  & acw      & 21.75    & 3486     & Hindi                                   & hin      & 10.93    & 6539     & Huaxcaleca Nahuatl               & nhq      & 5.94     & 1980     \\
Huba                           & hbb      & 6.14     & 2290     & Huitepec Mixtec                         & mxs      & 5.87     & 1570     & Hula                             & hul      & 6.10     & 2009     \\
Hungarian                      & hun      & 63.64    & 41688    & Hunjara-Kaina Ke                        & hkk      & 4.19     & 955      & Hwana                            & hwo      & 7.26     & 2405     \\
Icelandic                      & isl      & 2.13     & 732      & Idakho-Isukha-Tiriki                    & ida      & 6.28     & 2280     & Idoma                            & idu      & 6.90     & 2665     \\
Igbo                           & ibo      & 9.49     & 2190     & Igo                                     & ahl      & 6.73     & 2596     & Ikposo                           & kpo      & 5.67     & 1524     \\
Ikwere                         & ikw      & 6.03     & 1519     & Indonesian                              & ind      & 14.54    & 6955     & Interlingua                      & ina      & 4.73     & 4447     \\
Irish                          & gle      & 9.78     & 2812     & Isekiri                                 & its      & 7.89     & 2394     & Isoko                            & iso      & 6.15     & 1789     \\
Italian                        & ita      & 261.38   & 175120   & Ito                                     & itw      & 6.54     & 2301     & Itzá                             & itz      & 4.73     & 1003     \\
Ixtayutla Mixtec               & vmj      & 6.34     & 1767     & Izon                                    & ijc      & 5.84     & 1706     & Jambi Malay                      & jax      & 6.81     & 2335     \\
Japanese                       & jpn      & 24.65    & 16747    & Jaunsari                                & jns      & 4.42     & 909      & Javanese                         & jav      & 8.58     & 2409     \\
Jiba                           & juo      & 6.88     & 2637     & Jju                                     & kaj      & 5.90     & 2187     & Juxtlahuaca Mixtec               & vmc      & 5.93     & 1755     \\
Kabras                         & lkb      & 6.28     & 2027     & Kabuverdianu                            & kea      & 8.01     & 2135     & Kabyle                           & kab      & 144.06   & 152224   \\
Kachi Koli                     & gjk      & 4.46     & 1212     & Kairak                                  & ckr      & 6.47     & 2338     & Kalabari                         & ijn      & 6.86     & 2351     \\
Kalenjin                       & kln      & 13.37    & 11057    & Kamba (Kenya)                           & kam      & 10.94    & 2618     & Kamo                             & kcq      & 6.85     & 2783     \\
Kanauji                        & bjj      & 6.46     & 2290     & Kanembu                                 & kbl      & 6.69     & 2265     & Kannada                          & kan      & 6.00     & 1735     \\
Karekare                       & kai      & 7.09     & 2320     & Kashmiri                                & kas      & 4.94     & 989      & Kathoriya Tharu                  & tkt      & 6.84     & 2241     \\
Kazakh                         & kaz      & 9.68     & 3093     & Keiyo                                   & eyo      & 6.28     & 2000     & Khana                            & ogo      & 6.66     & 2102     \\
Khmer                          & khm      & 5.23     & 1282     & Kinga                                   & zga      & 9.67     & 3620     & Kinnauri                         & kfk      & 4.31     & 1304     \\
Kinyarwanda                    & kin      & 1309.82  & 929156   & Kirghiz                                 & kir      & 9.31     & 3967     & Kirya-Konz@l                     & fkk      & 5.86     & 2171     \\
Kochila Tharu                  & thq      & 6.82     & 2392     & Kohumono                                & bcs      & 7.10     & 2431     & Kok Borok                        & trp      & 7.88     & 2482     \\
Kol (Papua New Guinea)         & kol      & 6.14     & 2186     & Koma                                    & kmy      & 6.49     & 2431     & Konkani (individual language)    & knn      & 8.10     & 1802     \\
Konzo                          & koo      & 9.49     & 3118     & Korean                                  & kor      & 6.47     & 2154     & Korwa                            & kfp      & 7.08     & 1535     \\
Kota                           & kfe      & 9.50     & 3394     & Kuanua                                  & ksd      & 6.02     & 2064     & Kuanyama                         & kua      & 6.31     & 2269     \\
Kui (India)                    & uki      & 6.66     & 1680     & Kulung (Nigeria)                        & bbu      & 6.24     & 2285     & Kuot                             & kto      & 6.02     & 1740     \\
Kushi                          & kuh      & 6.75     & 2100     & Kwambi                                  & kwm      & 6.64     & 2519     & Lala-Roba                        & lla      & 6.31     & 2226     \\
Lamang                         & hia      & 7.12     & 2471     & Lao                                     & lao      & 5.68     & 1516     & Larike-Wakasihu                  & alo      & 6.12     & 1812     \\
Latgalian                      & ltg      & 5.50     & 4534     & Latvian                                 & lav      & 27.82    & 15805    & Levantine Arabic                 & apc      & 5.75     & 1886     \\
Liana-Seti                     & ste      & 6.21     & 1603     & Liberia Kpelle                          & xpe      & 6.21     & 2482     & Liberian English                 & lir      & 7.58     & 2239     \\
Libyan Arabic                  & ayl      & 16.25    & 2484     & Ligurian                                & lij      & 9.00     & 3460     & Lijili                           & mgi      & 6.79     & 2511     \\
Lingala                        & lin      & 11.32    & 2356     & Lithuanian                              & lit      & 19.02    & 10604    & Logooli                          & rag      & 5.93     & 2053     \\
Logudorese Sardinian           & src      & 6.53     & 2006     & Loloda                                  & loa      & 5.83     & 1272     & Longuda                          & lnu      & 6.59     & 2054     \\
Loxicha Zapotec                & ztp      & 6.29     & 2128     & Luo (Kenya and Tanzania)                & luo      & 12.18    & 5853     & Lushai                           & lus      & 10.02    & 3435     \\
Luxembourgish                  & ltz      & 6.24     & 1922     & Maasina Fulfulde                        & ffm      & 10.41    & 1664     & Maba                             & mde      & 8.80     & 2613     \\
Macedonian                     & mkd      & 7.58     & 3882     & Mafa                                    & maf      & 6.35     & 2312     & Malagasy, Southern Betsimisaraka & bzc      & 16.99    & 3293     \\
Malayalam                      & mal      & 8.73     & 3541     & Mali                                    & gcc      & 5.61     & 2367     & Malinaltepec Me'phaa             & tcf      & 6.21     & 2586     \\
Maltese                        & mlt      & 9.87     & 4172     & Mandara                                 & tbf      & 5.87     & 1562     & Mandarin Chinese                 & cmn      & 56.27    & 39265    \\
Mandjak                        & mfv      & 6.19     & 1920     & Manggarai                               & mqy      & 6.39     & 1702     & Mansoanka                        & msw      & 6.10     & 2317     \\
Maori                          & mri      & 11.67    & 2317     & Marathi                                 & mar      & 11.39    & 4158     & Marghi Central                   & mrt      & 6.30     & 2290     \\
Marghi South                   & mfm      & 6.57     & 2030     & Maria (India)                           & mrr      & 6.98     & 2403     & Masikoro Malagasy                & msh      & 13.89    & 3207     \\
Mazaltepec Zapotec             & zpy      & 6.10     & 1380     & Mazatlán Mazatec                        & vmz      & 5.15     & 1600     & Mazatlán Mixe                    & mzl      & 6.23     & 1603     \\
Mbe                            & mfo      & 6.50     & 2330     & Mekeo                                   & mek      & 5.29     & 1527     & Meru                             & mer      & 6.07     & 1878     \\
Mesopotamian Arabic            & acm      & 3.69     & 775      & Mewari                                  & mtr      & 1.88     & 374      & Mitlatongo Mixtec                & vmm      & 6.45     & 1482     \\
Miya                           & mkf      & 6.42     & 1930     & Modern Greek                            & ell      & 9.68     & 4417     & Moksha                           & mdf      & 0.26     & 175      \\
Mom Jango                      & ver      & 7.08     & 2372     & Mongolian                               & mon      & 11.04    & 4467     & Moroccan Arabic                  & ary      & 7.89     & 1658     \\
Motu                           & meu      & 6.65     & 2247     & Musi                                    & mui      & 6.58     & 2214     & Naba                             & mne      & 4.92     & 1529     \\
Najdi Arabic                   & ars      & 19.10    & 3094     & Nalik                                   & nal      & 6.07     & 1403     & Ndonga                           & ndo      & 5.92     & 2300     \\
Neapolitan                     & nap      & 6.30     & 2632     & Nepali (macrolanguage)                  & nep      & 8.74     & 2919     & Ngamo                            & nbh      & 5.72     & 1566     \\
Ngas                           & anc      & 6.69     & 2440     & Ngizim                                  & ngi      & 6.44     & 2342     & Nigerian Fulfulde                & fuv      & 5.60     & 1727     \\
Nimadi                         & noe      & 5.85     & 1971     & Nobiin                                  & fia      & 5.85     & 1595     & North Mesopotamian Arabic        & ayp      & 6.53     & 989      \\
North Moluccan Malay           & max      & 6.18     & 1575     & Northern Betsimisaraka Malagasy         & bmm      & 18.86    & 4052     & Northern Hindko                  & hno      & 4.51     & 1727     \\
Northern Kurdish               & kmr      & 5.84     & 5277     & Northern Pame                           & pmq      & 6.15     & 1417     & Northern Pashto                  & pbu      & 6.72     & 1023     \\
Northern Uzbek                 & uzn      & 11.84    & 2723     & Norwegian Bokmål                        & nob      & 8.15     & 2600     & Norwegian Nynorsk                & nno      & 0.50     & 464      \\
Notsi                          & ncf      & 4.80     & 1146     & Nyanja                                  & nya      & 7.92     & 2093     & Nyankpa                          & yes      & 6.38     & 1838     \\
Nzanyi                         & nja      & 5.96     & 1806     & Occitan                                 & oci      & 10.83    & 2909     & Od                               & odk      & 4.48     & 1261     \\
Odia                           & ory      & 5.05     & 2665     & Odual                                   & odu      & 6.17     & 1983     & Omani Arabic                     & acx      & 21.76    & 3408     \\
Orma                           & orc      & 11.12    & 2121     & Oromo                                   & orm      & 5.03     & 1350     & Ossetian                         & oss      & 0.66     & 414      \\
Pahari-Potwari                 & phr      & 5.82     & 2360     & Panjabi                                 & pan      & 6.13     & 2323     & Papuan Malay                     & pmy      & 6.52     & 2559     \\
Pedi                           & nso      & 6.73     & 1232     & Pero                                    & pip      & 5.45     & 1843     & Persian                          & fas      & 40.79    & 32225    \\
Petats                         & pex      & 6.42     & 2262     & Piemontese                              & pms      & 15.95    & 2790     & Piya-Kwonci                      & piy      & 7.03     & 1888     \\
Plateau Malagasy               & plt      & 19.05    & 4454     & Polish                                  & pol      & 41.69    & 25813    & Poqomam                          & poc      & 6.15     & 1699     \\
Portuguese                     & por      & 33.59    & 24643    & Pulaar                                  & fuc      & 14.29    & 2689     & Pular                            & fuf      & 13.69    & 3074     \\
Pökoot                         & pko      & 6.26     & 2278     & Qaqet                                   & byx      & 5.84     & 2043     & Quiotepec Chinantec              & chq      & 6.38     & 2127     \\
Rana Tharu                     & thr      & 5.17     & 1397     & Rangi                                   & lag      & 6.23     & 2247     & Rapoisi                          & kyx      & 6.16     & 2477     \\
Ratahan                        & rth      & 5.43     & 1142     & Rayón Zoque                             & zor      & 6.59     & 2064     & Romanian                         & ron      & 13.39    & 7429     \\
Romansh (Sursilvan)            & roh      & 2.49     & 1591     & Romansh (Vallader)                      & roh      & 1.00     & 557      & Rombo                            & rof      & 6.54     & 2286     \\
Rotokas                        & roo      & 6.03     & 1791     & Russian                                 & rus      & 44.40    & 28714    & Sacapulteco                      & quv      & 6.36     & 1918     \\
Saidi Arabic                   & aec      & 7.07     & 1922     & Sakalava Malagasy                       & skg      & 8.59     & 1841     & Saleman                          & sau      & 5.76     & 1547     \\
Samba Daka                     & ccg      & 6.20     & 1877     & Samba Leko                              & ndi      & 6.78     & 2473     & San Felipe Otlaltepec Popoloca   & pow      & 6.24     & 1490     \\
San Francisco Del Mar Huave    & hue      & 5.94     & 1542     & San Juan Atzingo Popoloca               & poe      & 5.86     & 1627     & San Martín Itunyoso Triqui       & trq      & 5.55     & 2022     \\
San Miguel El Grande Mixtec    & mig      & 6.36     & 1744     & Santa Catarina Albarradas Zapotec       & ztn      & 5.96     & 1748     & Santali (Ol Chiki)               & sat      & 0.44     & 333      \\
Saposa                         & sps      & 5.77     & 1822     & Saraiki                                 & skr      & 1.50     & 1556     & Sardinian                        & srd      & 1.16     & 923      \\
Saya                           & say      & 5.36     & 1552     & Serbian                                 & srp      & 9.51     & 4162     & Shona                            & sna      & 7.48     & 1924     \\
Siar-Lak                       & sjr      & 5.93     & 1538     & Sibe                                    & nco      & 6.31     & 1836     & Sicilian                         & scn      & 8.96     & 2421     \\
Sikkimese                      & sip      & 4.06     & 1085     & Sinaugoro                               & snc      & 6.33     & 1859     & Sindhi                           & snd      & 9.47     & 2903     \\
Sinhala                        & sin      & 8.07     & 2395     & Sinicahua Mixtec                        & xti      & 6.33     & 1650     & Sipacapense                      & qum      & 6.37     & 1850     \\
Siwai                          & siw      & 6.66     & 2386     & Slovak                                  & slk      & 13.48    & 8917     & Slovenian                        & slv      & 7.24     & 3414     \\
Solos                          & sol      & 6.41     & 2117     & Somali                                  & som      & 9.26     & 2362     & Soninke                          & snk      & 6.72     & 2760     \\
Southeastern Nochixtlán Mixtec & mxy      & 5.74     & 1276     & Southern Pashto                         & pbt      & 6.82     & 1052     & Soyaltepec Mazatec               & vmp      & 6.09     & 1613     \\
Spanish                        & spa      & 515.42   & 355892   & Standard Arabic                         & arb      & 37.31    & 30271    & Standard Estonian                & ekk      & 11.75    & 5323     \\
Standard Malay                 & zsm      & 7.28     & 2110     & Standard Moroccan Tamazight             & zgh      & 0.73     & 840      & Sudanese Arabic                  & apd      & 6.71     & 1674     \\
Sulka                          & sua      & 6.17     & 1769     & Swahili (individual language)           & swh      & 79.69    & 48919    & Swedish                          & swe      & 15.40    & 10011    \\
Tae'                           & rob      & 5.66     & 1452     & Tahaggart Tamahaq                       & thv      & 1.62     & 350      & Taita                            & dav      & 2.29     & 2096     \\
Tajik                          & tgk      & 6.66     & 1898     & Tamil                                   & tam      & 89.09    & 47680    & Tandroy-Mahafaly Malagasy        & tdx      & 1.71     & 460      \\
Tangale                        & tan      & 7.00     & 2617     & Tanosy Malagasy                         & txy      & 6.05     & 1632     & Tarok                            & yer      & 6.15     & 2067     \\
Tatar                          & tat      & 8.84     & 8394     & Tedaga                                  & tuq      & 6.24     & 1953     & Telugu                           & tel      & 5.82     & 1780     \\
Teop                           & tio      & 6.01     & 1795     & Tepinapa Chinantec                      & cte      & 6.64     & 2136     & Tera                             & ttr      & 6.39     & 2213     \\
Terei                          & buo      & 6.52     & 2065     & Termanu                                 & twu      & 6.36     & 1516     & Tesaka Malagasy                  & tkg      & 12.96    & 2700     \\
Tetelcingo Nahuatl             & nhg      & 6.25     & 1865     & Thai                                    & tha      & 43.69    & 34889    & Tidaá Mixtec                     & mtx      & 5.73     & 1327     \\
Tidore                         & tvo      & 6.85     & 1459     & Tigak                                   & tgc      & 5.77     & 1442     & Tigre                            & tig      & 2.94     & 1972     \\
Tigrinya                       & tir      & 0.03     & 20       & Tilquiapan Zapotec                      & zts      & 6.03     & 1689     & Tinputz                          & tpz      & 5.03     & 1462     \\
Tlacoapa Me'phaa               & tpl      & 6.36     & 2811     & Tlacoatzintepec Chinantec               & ctl      & 6.08     & 1979     & Toki Pona                        & tok      & 2.55     & 2629     \\
Tomoip                         & tqp      & 5.69     & 1612     & Tondano                                 & tdn      & 5.12     & 1201     & Tonsea                           & txs      & 6.34     & 1584     \\
Tooro                          & ttj      & 6.68     & 1976     & Torau                                   & ttu      & 5.44     & 2028     & Tsimihety Malagasy               & xmw      & 11.37    & 2664     \\
Tsotso                         & lto      & 6.19     & 2251     & Tswana                                  & tsn      & 1.31     & 1078     & Tugen                            & tuy      & 6.68     & 1986     \\
Tula                           & tul      & 5.63     & 1925     & Tulu                                    & tcy      & 6.93     & 2172     & Tungag                           & lcm      & 5.93     & 1933     \\
Tunisian Arabic                & aeb      & 21.14    & 2986     & Turkana                                 & tuv      & 5.69     & 2116     & Turkish                          & tur      & 49.46    & 41561    \\
Turkmen                        & tuk      & 1.03     & 739      & Tututepec Mixtec                        & mtu      & 6.18     & 1658     & Ubaghara                         & byc      & 7.05     & 2663     \\
Uighur                         & uig      & 166.71   & 107645   & Ukrainian                               & ukr      & 38.30    & 28879    & Umbundu                          & umb      & 5.75     & 1034     \\
Upper Sorbian                  & hsb      & 1.48     & 808      & Urdu                                    & urd      & 13.78    & 8982     & Uzbek                            & uzb      & 61.39    & 49656    \\
Vai                            & vai      & 5.64     & 1991     & Vietnamese                              & vie      & 9.35     & 4398     & Votic                            & vot      & 0.11     & 96       \\
Võro                           & vro      & 11.43    & 2266     & Waci Gbe                                & wci      & 6.22     & 2223     & Wadiyara Koli                    & kxp      & 5.02     & 1476     \\
Waja                           & wja      & 6.72     & 2135     & Wanga                                   & lwg      & 6.07     & 2236     & Wapan                            & juk      & 6.10     & 1983     \\
Warji                          & wji      & 7.94     & 2708     & Welsh                                   & cym      & 20.19    & 10269    & Wemale                           & weo      & 6.05     & 1381     \\
Western Frisian                & fry      & 5.53     & 3924     & Western Juxtlahuaca Mixtec              & jmx      & 6.50     & 2378     & Western Maninkakan               & mlq      & 5.76     & 2038     \\
Western Mari                   & mrj      & 14.89    & 14325    & Western Niger Fulfulde                  & fuh      & 6.25     & 1176     & Western Panjabi                  & pnb      & 6.52     & 1855     \\
Wolof                          & wol      & 7.41     & 2013     & Xanaguía Zapotec                        & ztg      & 5.75     & 1620     & Xhosa                            & xho      & 9.66     & 2648     \\
Yace                           & ekr      & 7.20     & 2613     & Yakut                                   & sah      & 3.76     & 2195     & Yalahatan                        & jal      & 7.93     & 2207     \\
Yekhee                         & ets      & 6.49     & 1808     & Yoruba                                  & yor      & 10.54    & 3388     & Yue Chinese                      & yue      & 23.26    & 17377    \\
Yutanduchi Mixtec              & mab      & 6.68     & 1951     & Zacatlán-Ahuacatlán-Tepetzintla Nahuatl & nhi      & 0.03     & 23       & Zarma                            & dje      & 7.50     & 3244     \\
Zaza                           & zza      & 0.79     & 734      & Zulu                                    & zul      & 10.28    & 2126     & Ömie                             & aom      & 4.36     & 996   \\

\end{supertabular}

\vspace{1ex}
\captionof{table}{Detailed configuration of the pretraining dataset. Duration is measured in hours, and sentences denote the number of speech–text pairs. Language names follow the Omnilingual ASR configuration, and ISO codes follow ISO-639-3.}
\label{tab:appendix_config}

\twocolumn 
\endgroup

\end{document}